\documentclass[letterpaper, 10 pt, conference]{ieeeconf}  

\IEEEoverridecommandlockouts                              

\usepackage{amsmath,amssymb,amsfonts}
\usepackage{algorithmic}

\usepackage{textcomp}
\usepackage{xcolor}
\usepackage{subfig}
\usepackage{multirow}
\usepackage{booktabs}
\usepackage{makecell}
\usepackage{balance}
\usepackage{caption}
\usepackage{url}
\usepackage{cite}
\usepackage{graphicx}
\usepackage{subfloat}

\title{\LARGE \bf
Robopheus: A Virtual-Physical Interactive Mobile Robotic Testbed 
}

\author{Xuda Ding$^{1}$, Han Wang$^{2}$, Hongbo Li$^{3}$, Hao Jiang$^{1}$ and Jianping He$^{1}$
\thanks{$^{1}$The authors are with Department of Automation, Shanghai Jiao Tong University, Shanghai, 200240,  China. E-mails: {\tt\small \{dingxuda, mouse826612011, jphe\}@sjtu.edu.cn}}
\thanks{$^{2}$Han Wang is with the Department of Engineering Science, University of Oxford, Oxford, UK.
        E-mail: {\tt\small han.wang@linacre.ox.ac.uk}}%
\thanks{$^{3}$Hongbo Li is with the Pillar of Engineering Systems and Design, Singapore University of Technology and Design, Singapore 487372. E-mail: {\tt\small hongbo\_li@mymail.sutd.edu.sg}}%
}

\begin{document}

\maketitle
\thispagestyle{empty}
\pagestyle{empty}

\begin{abstract}
The mobile robotic testbed is an essential and critical support to verify the effectiveness of mobile robotics research. 
This paper introduces a novel multi-robot testbed, named Robopheus, which exploits the ideas of virtual-physical modeling in digital-twin. Unlike most existing testbeds, the developed Robopheus constructs a bridge that connects the traditional physical hardware and virtual simulation testbeds, providing scalable, interactive, and high-fidelity simulations-tests on both sides. 
Another salient feature of the Robopheus is that it enables a new form to learn the actual models from the physical environment dynamically and is compatible with heterogeneous robot chassis and controllers. 
In turn, the virtual world's learned models are further leveraged to approximate the robot dynamics online on the physical side. 
Extensive experiments demonstrate the extraordinary performance of the Robopheus. 
Significantly, the physical-virtual interaction design increases the trajectory accuracy of a real robot by 300$\%$, 
compared with that of not using the interaction.

\end{abstract}

\section{INTRODUCTION}
The study of mobile robotics has received considerable attention in the last decade and has been deployed in numerous industrial and military applications. 
Fruitful tools have been adopted to develop the research, such as distributed control \cite{dimarogonas2011distributed,shamma2008cooperative}, learning \cite{bucsoniu2010multi,vinyals2019grandmaster} and information  theories \cite{such2014survey}. 
Specifically, due to the rapid development of computer and mechatronics technologies, mobile robotic testbeds have emerged as a critical part of the research cycle to validate the theoretical results and protocols \cite{rohmer2013v,freese2010virtual,lee2018distributed, pickem2015gritsbot,paull2017duckietown,wu2019phoenix,pickem2017robotarium, wilson2020robotarium,liang2018plugo}.


Generally, testbeds can be divided into two categories: virtual and physical. 
The virtual testbed enjoys customized experiment fashion, convenient maintenance, and flexible scalability, applying well to the illustration of various theoretical validation \cite{rohmer2013v, freese2010virtual, lee2018distributed,vladareanu2015optimization,carlsson1993dive}. 
The major drawback is that the real world's typical implementation issues are usually simplified or even omitted in the virtual environment (e.g., wheel slip, friction, computation time, and actuator constraints), failing to fulfill a high-fidelity requirement. On the contrary, the physical testbed provides a way to practically validate the proposed theories' effectiveness, help improve the theory inadequacy, and discover new issues in real implementation \cite{pickem2015gritsbot,paull2017duckietown,pickem2017robotarium,wu2019phoenix,wilson2020robotarium}. 
A representative example is Robotarium  \cite{pickem2017robotarium, wilson2020robotarium}, which provides worldwide remote access for researchers to use the testbed for experiments. 
Nevertheless, physical testbeds are challenging to apply for various scenarios due to the limited space and resources. 
Besides, the robots provided by the physical testbed may not meet the need of various control strategies, due to the fixed robot kinetics type.
\begin{figure}[t]
	\centering
	\includegraphics[width=1\linewidth]{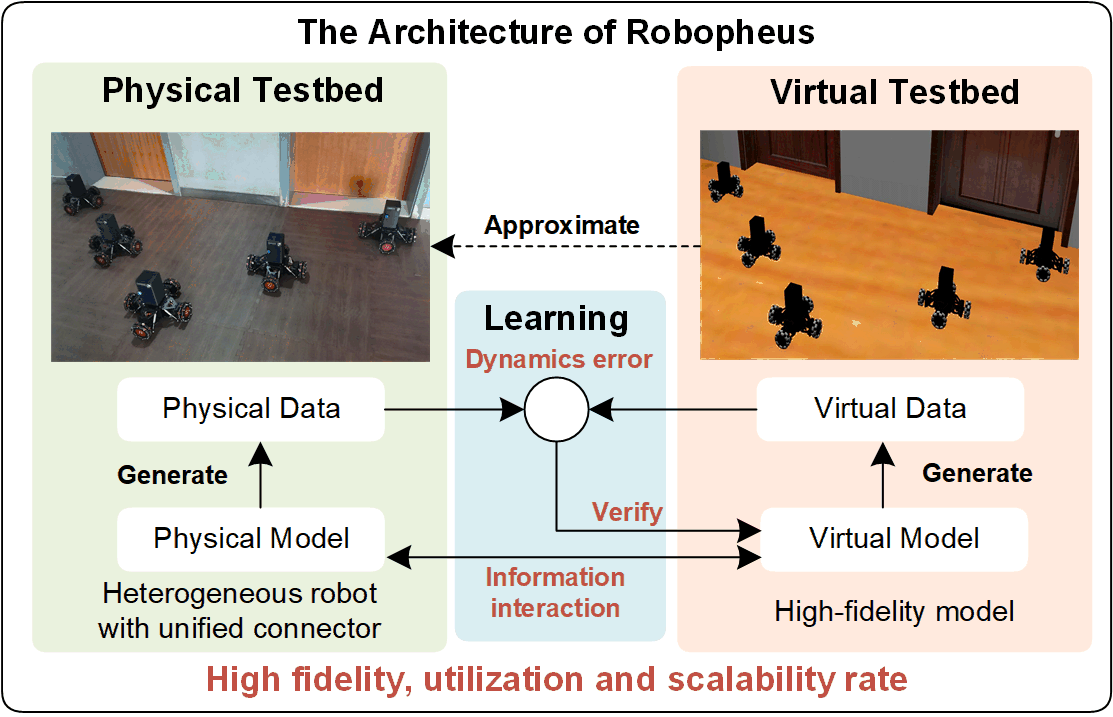}
	\caption{The architecture of Robopheus}
	\label{fig:screenshot001}
	\vspace{5 pt}
\end{figure}

This paper presents a novel testbed that combines the merits of both traditional virtual testbed and the physical testbed and even beyond can be achieved. On the one hand, the state of art digital-twin idea provides the potential for virtual testbeds to perform high-fidelity simulations based on accurate digital models of physical objects \cite{boschert2016digital, tao2019make}. 
On the other hand, some prior works have shown the possibility to identify the internal system structure and parameters from accessible data \cite{carleo2017solving,liu2019dynamic,zhao2014support}. 
However, the real-time interaction between the physical and virtual worlds is blocked, and the two parts run independently. 
Besides, it is tough to construct heterogeneous object models that consider the practical factors (e.g., wheel slip and friction).

Therefore, we aim to build a new physical-virtual interaction based mobile robotic testbed, named Robopheus, which provides high-fidelity simulations in the virtual world and high-accuracy tests in physical environments. 
The physical part includes our self-designed heterogeneous robots and supports the real-time model construction and calibration in the virtual part. The virtual part provides extensive simulation settings and high-fidelity objective models, including the accessed environments and heterogeneous robots to all users. The testbed is compatible with heterogeneous kinematics robots and controllers, and users can install the virtual testbed on their computers. 
Specifically, the robot and environment models can be learned from the robot state evolution in the physical testbed. 
Furthermore, the online learned models in the virtual environment can be used for predicting the future states of robots, providing an optimal strategy for actual control in physical testbed. 
The architecture of the Robopheus is shown in Fig. \ref{fig:screenshot001}).

The achieve the Robopheus, the main challenges lie in three aspects. 
i) Due to the variety of heterogeneous kinematics robot chassis and the plug and play controller design. It is challenging to construct a unified connector that is compatible with heterogeneous robots.
ii) The environment factors such as wheel slip, friction, and actuator constraints are not prior knowledge in the virtual world, making it hard to construct the model structure and high-fidelity. 
iii) As the Robopheus is combined with various small and coupled systems (e.g., positioning system, robot-embedded control system, and communication systems), the global stability of robot control remains unsolved. 
Our work overcame these challenges, and the contributions are as follows. 

\begin{itemize}
	\item To the best of our knowledge, it is the first time to develop an interactive virtual-physical mobile robotic testbed. The constructed bridge of two sides overcomes the disadvantages of a single physical or virtual testbed and provides a newly interactive-learning form to enhance the real-time task performance of actual robots. 
	\item We self-design heterogeneous kinematics robot chassis and controllers, along with their unified connector. 
	The design covers more application scenarios and requirements for algorithm verification, thus improves the scalability and accessibility of the physical testbed.
	\item We propose a real-time method to learn the practical dynamic models of robots from the real operation data, considering wheel slip, friction, and actuator constraints. The online learned models provide real-time feedback to improve the high-fidelity of the virtual simulation and can be utilized for motion predictions. 
\end{itemize}

The remainder of this paper is organized as follows.
Section \uppercase\expandafter{\romannumeral2} gives descriptions of physical-virtual digital-twin testbed system.
Section \uppercase\expandafter{\romannumeral3} provides the design of the physical testbed and details of the heterogeneous kinematics robot chassis, controllers, and the connector.
Section \uppercase\expandafter{\romannumeral4} provides the design of the virtual testbed and details of models.
Section \uppercase\expandafter{\romannumeral5} shows the physical-virtual testbed and the learning procedure and effect of the physical-virtual interaction.
Lastly, conclusions are given in section \uppercase\expandafter{\romannumeral6}.

\section{ROBOPHEUS OVERVIEW}

The Robopheus is constituted by physical testbed, virtual testbed, and learning procedure (see Fig. \ref{fig:screenshot001}). Currently, the interaction between physical and virtual testbeds is the kinetics data and model parameters. 
Models of physical and virtual testbed generate data separately, and the data interaction triggers the model learning procedure and drive the virtual model to approximate the actual physical model. Based on the above mechanism, the Robopheus provides high fidelity, high utilization, and generalization rate for robot simulations, which will be demonstrated in the experimental part.

\begin{figure}[t]
	\centering
	\includegraphics[width=1\linewidth]{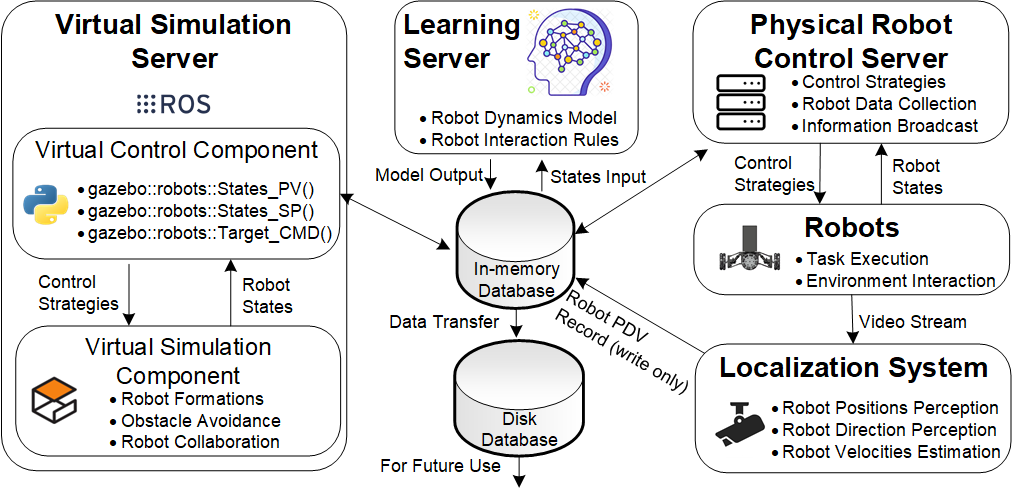}
	\caption{The structure of Robopheus}
	\label{fig:screenshot002}
	\vspace{-5 pt}
\end{figure}
The Robopheus is constituted by virtual simulation sever, learning server and physical robot control server (see Fig. \ref{fig:screenshot002}). Besides servers, system has physical robots for task executions and localization sub-system (details in Section III.A) for robot pose detection. Real time data is aggregated into the in-memory database before being transferred into disk database, for ensuring low latency communication and independent operation of each part.

In the system, the Physical control server communicates with robots via wireless protocol (e.g., ZigBee, Digimesh, and WIFI), controls robot movements, and requests robot states. The data is transferred into the in-memory database via wired TCP/IP protocol.
The virtual simulation server is with virtual control and simulation components and performs a virtual testbed. The control component is a digital twin of the physical robot control server, built by Python, releases topics to provide control strategies, and requests robot states in the simulation component. Topics are written in the form of standard Gazebo statement, for the simulation component is built in Gazebo. 
The simulation component accomplishes robot simulations in a virtual environment, such as robot formations, obstacle avoidance, and robot collaborations. 
The data produced by the virtual simulation server is transferred into the in-memory database directly via wired TCP/IP protocol. 
The learning server uses data of virtual and physical testbeds from the in-memory database and learns information such as robot dynamics models and interaction rules. The output data of the server is back to the in-memory database.
Servers are deployed into different operating systems (OS), such as the Robot Operating System (ROS) under UBUNTU 16.04 and Windows 10 OS. For operational convenience, we choose VMware ESXi 7.0 for server management \cite{ESXI}. The virtual switch in ESXi is on the trunk mode.

\section{PHYSICAL TESTBED DESIGN}

This section gives details of the physical part of Robopheust, including localization and control subsystem design, heterogeneous robots and controllers design, and robot control method.

\subsection{Localization and Control Subsystem}

To detect poses of robots, a high-speed camera array with a 120 Hz frequency is installed under the roof. The array can cover a large-scale area of the physical testbed and can be expanded. The top of the robot has a quick-respond (QR) code picture with identity information. The cameras capture QR codes to identify robots and their positions, directions. Further, velocities are estimated based on two adjacent samples and sampling intervals. 

\begin{figure}[t]
	\centering
	\includegraphics[width=1\linewidth]{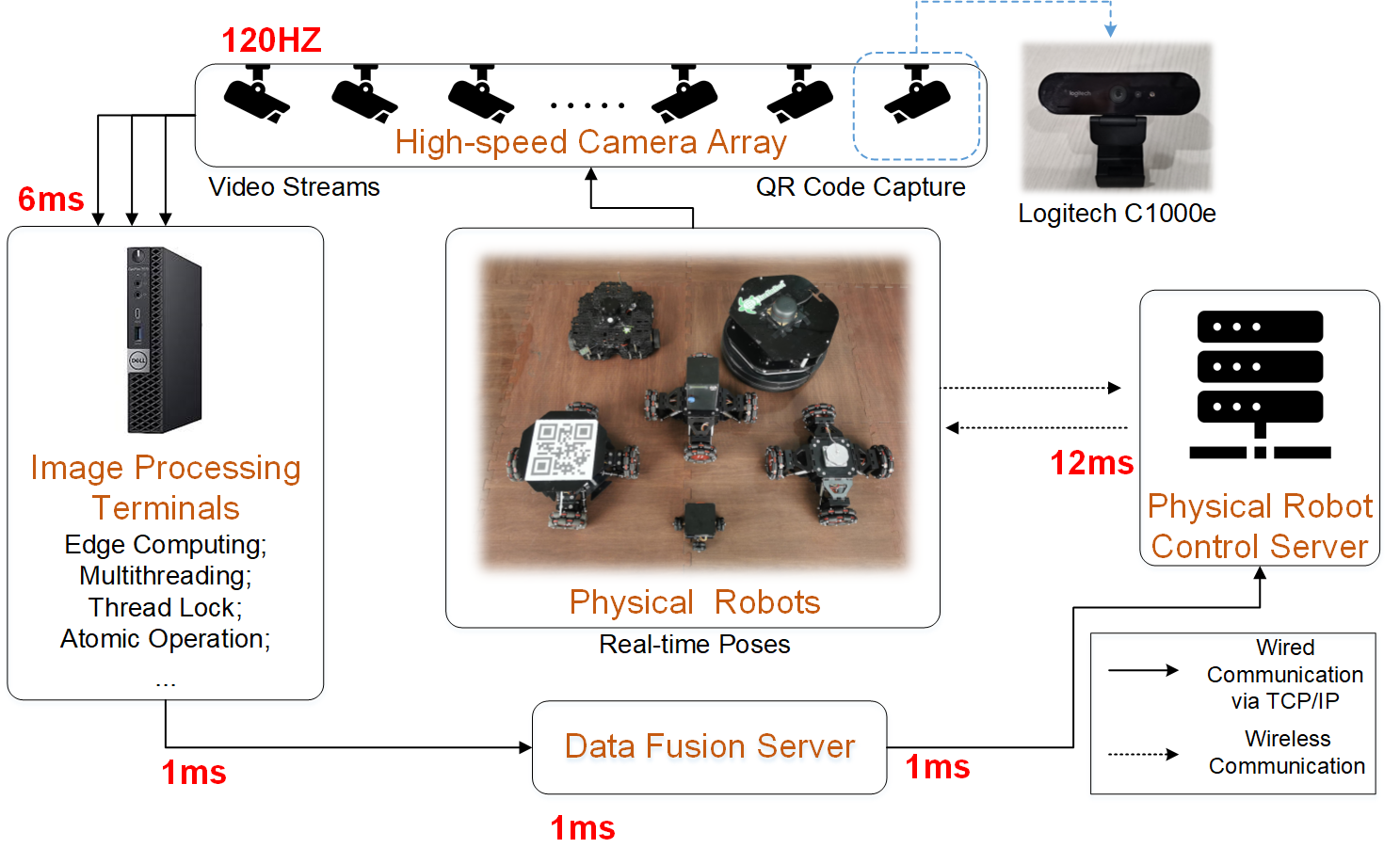}
	\caption{Localization and control subsystem structure}
	\label{fig:screenshot003}
\end{figure}

The 120 Hz video streams with images of 640 $\times$ 480 pixels (Logi C1000e \cite{C1000e}), are processed by image processing terminals (OptiPlex 7070 Micro Desktop \cite{7070}). Lens distortion correction is the first step of the processing, based on camera calibration model parameters. Then, image-enhancing procedure (e.g., optical flow, motion segmentation, and tracking). Finally, the positions and angles of QR codes are detected.
To guarantee the processing speed, each terminal calculates the stream of one camera under multi-thread mode. This edge-computing embedded method keeps the computational time below 6 ms. Operations in processing programs, such as thread locks and atomic operations, ensure the information following the first-in-first-out rule. Data fusion server collects data from all terminals with 1 ms communication latency, converts the positions and angles of each camera coordinate to information of unified coordinate. 

Then, robot control algorithms, such as consensus algorithm \cite{olfati2004consensus}, leader-follower control \cite{listmann2009consensus} and leaderless control \cite{ren2007information,meng2011leaderless}, can be uploaded into physical robot control server. The server calculates the command output (e.g., target position and speed) and broadcasts the output to robots via wireless communication. In the meantime, the robot can send the message, such as encoder, inertial measurement unit (IMU), and magnetic-sensor feedback, back to the server. The message is used to achieve delicate control or data storage for further research. The one-way transmission latency of wireless communication is no more than 12 ms based on a hexadecimal short coding method. Overall, the control latency from QR code capture to robot execution is about 21 ms. Therefore, the total latency is about 25 ms with our computing and transmission resources. The sampling frequency is 120 HZ, which is enough for accurate low-speed, e.g., 2 cm/s, robot control (less than 1 \% error).

\subsection{Heterogeneous Robots and Controllers}

\begin{figure}
		\vspace{10pt}
		\centering
		\includegraphics[width=0.8\linewidth]{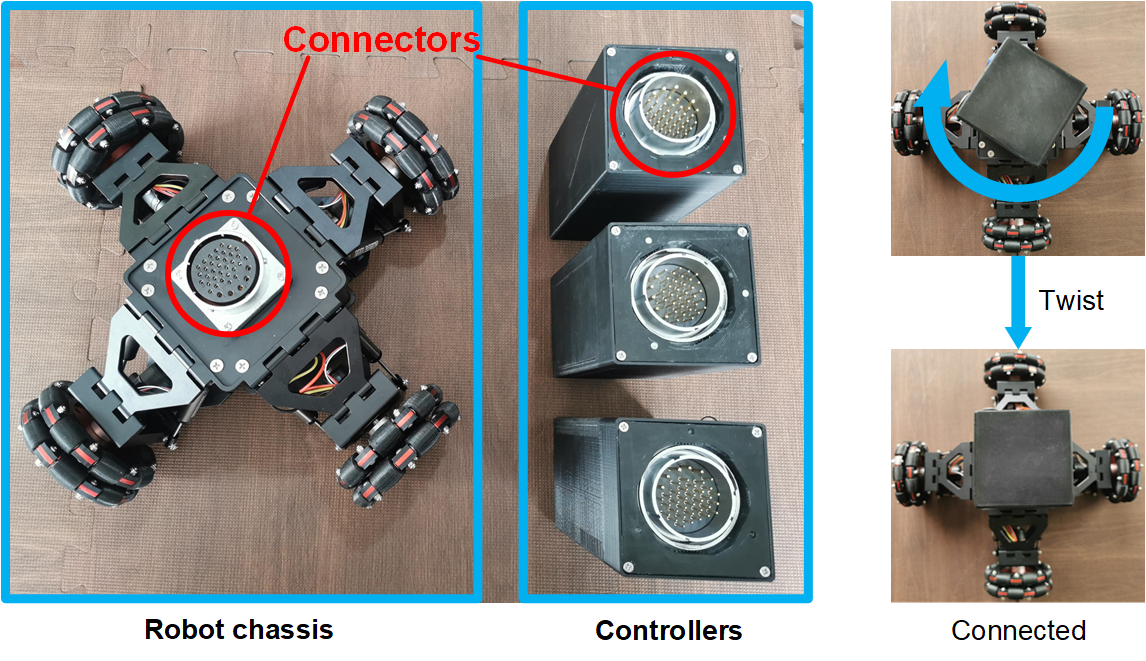}
		\caption{Connectors and install illustration}
		\label{fig:pic4}
	\end{figure}

To cover more application scenarios and requirements for algorithm verification, heterogeneous robots with various chassis and controllers that meet different performance are necessary. 
Currently, heterogeneous robot chassis with 6 different drive methods and controllers with 3 different processors are designed. Further, a unified connector for robot chassis and controllers is developed (shown in Fig. \ref{fig:pic4}), which provides an easy way to install or uninstall controllers on robots, since the connector is designed to be fool-proofing. 
		\begin{table}[t]
		\renewcommand\arraystretch{1.25}
		\vspace{5pt}
		\caption{Actuators and numbers of pins of different drive methods for heterogeneous robots}
		\vspace{-5pt}
		\begin{center}
			\begin{tabular}{llll}
				\toprule[1.5pt]
				\multicolumn{1}{c}{Drive method}  & \multicolumn{1}{c}{Omni-directional} & \multicolumn{1}{c}{2DD}  & \multicolumn{1}{c}{4DD}\\
				\midrule[0.75pt]
				\multicolumn{1}{c}{Actuators} & \multicolumn{1}{c}{4 motors}& \multicolumn{1}{c}{2 motors} &\multicolumn{1}{c}{4 motors} \\
				\multicolumn{1}{c}{Number of pins}& \multicolumn{1}{c}{28} & \multicolumn{1}{c}{14} & \multicolumn{1}{c}{28}\\
				\midrule[1.5pt]
				\multicolumn{1}{c}{Drive method}  & \multicolumn{1}{c}{FWD} & \multicolumn{1}{c}{RWD}  & \multicolumn{1}{c}{4WD}\\
				\midrule[0.75pt]
				\multirowcell{2}{Actuators} & \multicolumn{1}{c}{2 motors}& \multicolumn{1}{c}{2 motors} &\multicolumn{1}{c}{4 motors} \\
				& \multicolumn{1}{c}{1 steering gear}& \multicolumn{1}{c}{1 steering gear} &\multicolumn{1}{c}{1 steering gear} \\
				\multicolumn{1}{c}{Number of pins}& \multicolumn{1}{c}{17} & \multicolumn{1}{c}{17} & \multicolumn{1}{c}{31}\\
				\bottomrule[1.5pt]
			\end{tabular}
			\label{tab1}
		\end{center}
		\vspace{0pt}
	\end{table}
	
	A small Omni-directional robot chassis with 4 Omni wheels is developed in Fig. \ref{fig:pic4} (350 mm length, 350mm width, and 300 mm height). 
	4 brushless geared motors with a maximum power of 176 W are installed to drive the robot and carry more peripherals (up to 7 Kg). 
	4 independent suspension systems are designed to keep wheels on the floor all the time and guarantee the Omni-directional movement's stability. 
	Further, the robot can also achieve 2-wheel differential drive by setting 2 wheels into a passive mode (i.e., with no power supply). 
    4 more types of robot chassis are also designed, as the front-wheel-drive (FWD), rear-wheel-drive (RWD), four-wheel-drive (4WD), and 2 $\times$ 2 differential drive. 
	The actuators and numbers of each type of chassis's pins are listed in Tab. \ref{tab1}\footnote{Each motor has 3 pins for driven power, 4 pins for feedback communication. 
	Each steering gear has 2 pins for driven power, 1 pin for control signal}. 
	As for controllers, 8-bit (ATmega 2560), 32-bit (ATSAM 3X), and Nvidia Jetson TX2 are used as processors to meet different computing needs. 
	8-bit processors can achieve simple centralized control tasks, such as formation transforming, tracking, and covering. 
	32-bit processors with localization devices, such as ultra-wideband positioning, lidar, and global navigation satellite system (GNSS), can achieve self-localization and distributed control tasks. 
	Nvidia Jetson TX2 enables robots with powerful computational ability to realize intelligent self-control tasks with peripherals (e.g., depth cameras, lidar, and real-time kinematic based GNSS).
	
	Based on the number of pins in Tab. \ref{tab1}, we designed a 38-pin unified connector for robot chassis and controller connection (see in Fig. \ref{fig:pic4}). 4 pins are used for battery power supply and charging. 31 pins are used for 4 motors and 1 steering gear, which can meet all requirements for all drive methods in Tab. \ref{tab1}. The connector firmly links chassis and controller by a 60-degree twist, which is convenient for changing the chassis-controller pair.

\subsection{ROBOT CONTROL METHOD}

\begin{figure}[t]
		\centering
		\includegraphics[width=1\linewidth]{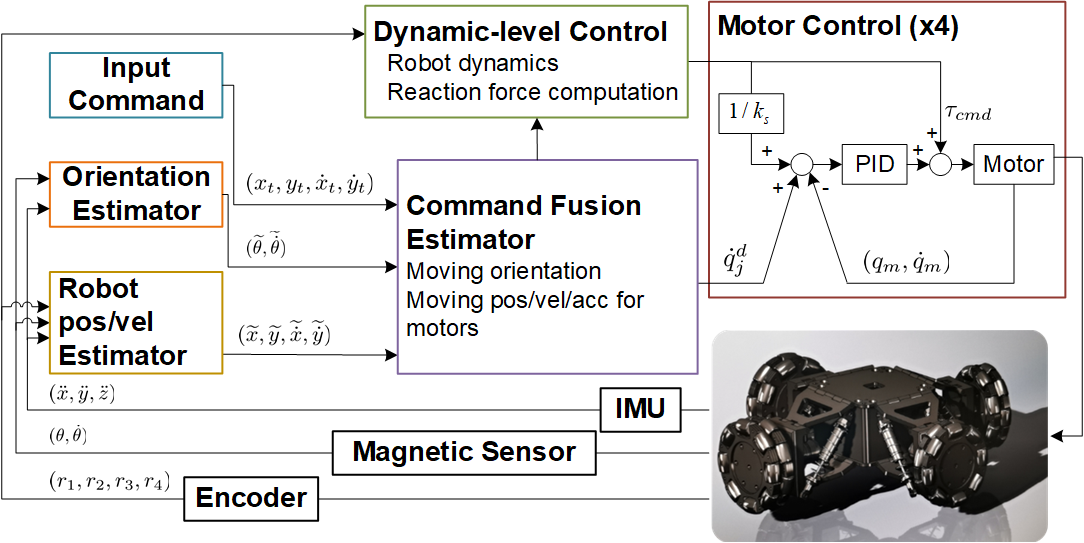}
		\caption{Robot control architecture}
		\label{fig:pic5}
	\end{figure}

Heterogeneous robots usually have different chassis and drive methods. In this subsection, we briefly introduce the robot control method for the Omni-directional robot as an example (see in Fig. \ref{fig:pic5}). The rest robots have a difference in dynamics, but their control architectures are the same.
Our proposed control architecture consists of three main components: 
i) a command fusion estimator which gives the command to motors based on estimated orientation, position, and velocity of the robot; 
ii) a dynamic-level control which gives the command to motors based on estimated velocity and acceleration of the robot; 
iii) a motor control which drives the robot based on commands. 
Together, they achieve a stable movement of the robot by coordinating the control of various actuators.
	
Specifically, accelerations at Cartesian coordinates ($\ddot{x},\ddot{y},\ddot{z}$), angle and angular velocity ($\theta, (\dot{\theta}))$,are measured by inertial measurement unit (IMU) and magnetic sensor (M-sensor) on the controller, respectively. The rotational speed of each wheel ($r_1, r_2, r_3, r_4$) is measured by encoder which is integrated in the motor, sent to the controller via unified connector. Robot position and velocity estimator calculates estimated position of the robot ($\tilde{x}, \tilde{y}, \tilde{\dot{x}}, \tilde{\dot{y}}$) based on the feedback of IMU, M-sensor and encoder. Orientation estimator calculates the estimated angle ($\tilde{\theta}$) and angular velocity ($\tilde{\dot{\theta}}$) based on the feedback of IMU and M-sensor. The input commands include target position ($x_t, y_t$) and target speed ($\dot{x}_t, \dot{y}_t$). Command fusion estimator gives command ($q_i^c$) to $i$th motor control and estimated moving states to dynamic-level control at the same time. Dynamic-level control uses the rotational speeds of wheels and estimated moving states to give command ($q_i^d$) to $i$th motor control based on structure and dynamic of the robot. The main algorithm in motor control is proportional integral derivative (PID). The current of $i$th motor ($q_i^m$) is considered as feedback of PID controller. Finally, motor controls drive the robot.

\section{VIRTUAL TESTBED DESIGN}
This section shows the virtual structure of Robopheus, including the simulation environment, components of robot mold, and system model of the whole virtual testbed.
\subsection{3D model}
We build our virtual testbed on physical engine simulation software Gazebo to realize high extensibility scenarios and high fidelity models of the real world. Gazebo integrated the ODE physics engine, OpenGL rendering, and support code for sensor simulation and actuator control. The virtual robot is constructed accurately according to the real one, which means that the materials, size, weight are all the same. In addition, we can also simulate many real-world elements that will influence robots' motion, e.g., ground friction, gravity, surface contact. 

\begin{figure}[b]
    \centering
    \includegraphics[width=1\linewidth]{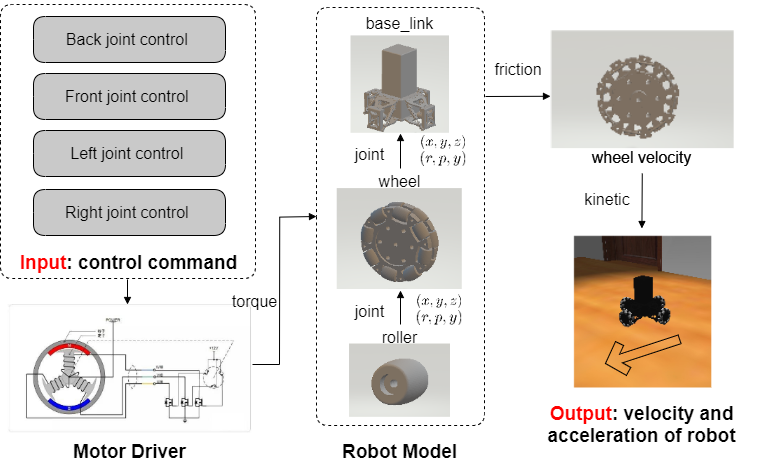}
    \caption{Virtual testbed structure on Gazebo}
    \label{fig:gazebo}
\end{figure}

To establish a close loop between the virtual and real systems, we have to build every part of the virtual model similar to the real one.  Unlike existing work on gazebo simulation\cite{rivera2019unmanned}, we not only one-to-one duplicate our robot mold on Gazebo, but also simulate the robot dynamic on it. The relationship among parts of the virtual structure can be found in Fig. \ref{fig:gazebo}.
The component of robot mold has three main parts:  roller, wheel, and base link. These three parts are connected by a virtual connector called a joint. 

The control input of the system include four parts, related to four joint linked base link and four wheels. In Gazebo, they are realized through the topic publisher. The first part of the system model is the motor driver, which will convert the input control signal to torque. Next, torque will drive the joint to rotate together with the wheel against ground friction. The output of the system is robot velocity and acceleration, which are both three dimensional.

\subsection{Dynamics Model Learning}
In this part, we implemented an online revision of the Robopheus virtual  model to continuously approach the dynamics of the robot in the real physical scene during the operation.
Although we have built every detail of the robot in a virtual environment, it is hard to get the dynamics of the robot for the complexity of the physical process. There are inevitably some differences between the virtual model and the physical robot, e.g., motor parameters. It is also vital to consider the impact of ground material and the accompanying phenomena (e.g., friction and slipping) on the robot dynamics in different physical scenarios.
Specific to the above problem and realizing a close loop between physical-virtual testbed, designing a real-time model learning method from the physical scenario to the virtual model is necessary. 

\begin{figure}[t]
    \centering
    \includegraphics[width=1\linewidth]{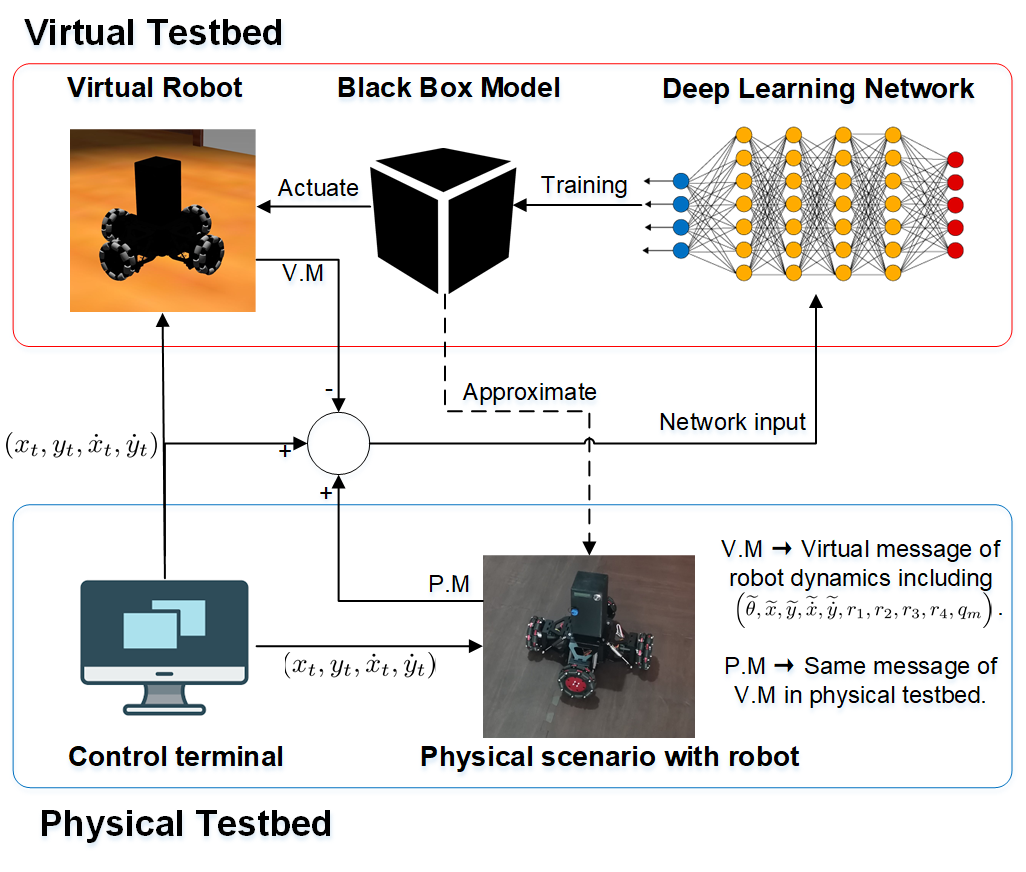}
    \caption{Dynamics model learning architecture}
    \label{fig:pic6}
    \vspace{-10 pt}
\end{figure}

Modern deep learning methods give an excellent solution to this question\cite{He_2016_CVPR}. We collect the input and output data for the dynamic model learning process. Details of this process can be seen in Fig. \ref{fig:screenshot002}, the learning server will get states input and calculate model, then transmit to the in-memory database. 
In the specific implementation, shown in Fig. \ref{fig:pic6}, we construct the robot model constrained by the physical environment as a black-box model and obtain the robot dynamic parameters of the physical testbed in real-time during the operation.
The black-box model is continuously approached to the robot dynamics in the physical testbed through the deep learning network. Finally, a high degree of fit between the virtual model and the actual physical model is achieved.

Through the above mechanism, the Robopheus can modify the virtual model online during the physical testbed operation, so that the virtual testbed realize a high-fidelity simulation of the actual scene (e.g., considering friction and slip). Then, using the revised model, we can make the optimization on the control strategy for the physical testbed to improve the real performance.

\section{EXPERIMENT EVALUATION ON VIRTUAL-PHYSICAL TESTBED}
In this section, we implement comparative experiments on the physical testbed and virtual testbed, respectively, to further illustrate the effectiveness of the Robopheus. The experiment consists of three parts: i) experiment on physical testbed without optimization from virtual testbed, on the other side the virtual testbed also gets no information from physical testbed to modify the model; ii) virtual testbed utilizes data from physical testbed, and then compute dynamics model learning process mentioned in section IV, subsection B to train the black-box model of virtual testbed; iii) in addition to the last experiment, compute the online supervising process to optimize the control protocol for the physical testbed. 

To show the utility of Robopheus, we choose the most basic as well as the typical task of the mobile robot, path planning. We set four targets for the robots, which are located on the four corners of a rectangle. Robots are desired to move precisely according to the desired trajectory towards the target. Unfortunately, this seemingly "simple" task is always hard to realize because of different noises in the physical environment, including ground friction, wheel slipping, and so on. 
	\begin{figure}[t]
	    \vspace{10pt}
		\centering
		\subfloat[trajectory in physical part]{
			\includegraphics[width=0.45\linewidth]{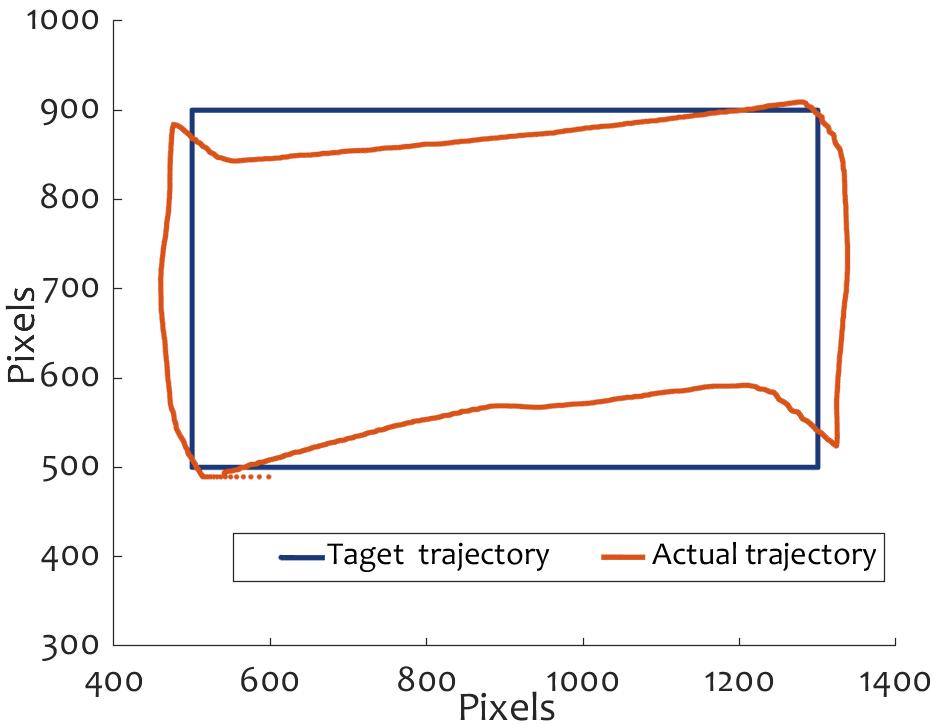}
		}
		\hspace{-5pt}
		\subfloat[trajectory in virtual part]{
			\includegraphics[width=0.45\linewidth]{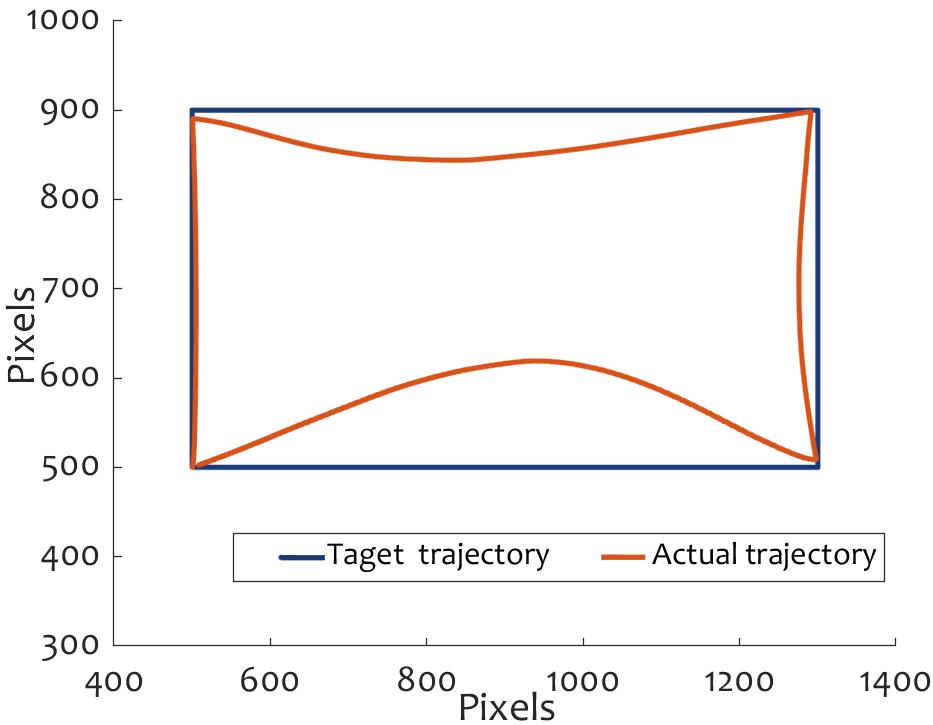}
		}
		
		\subfloat[trajectory error in physical part]{
			\includegraphics[width=0.45\linewidth]{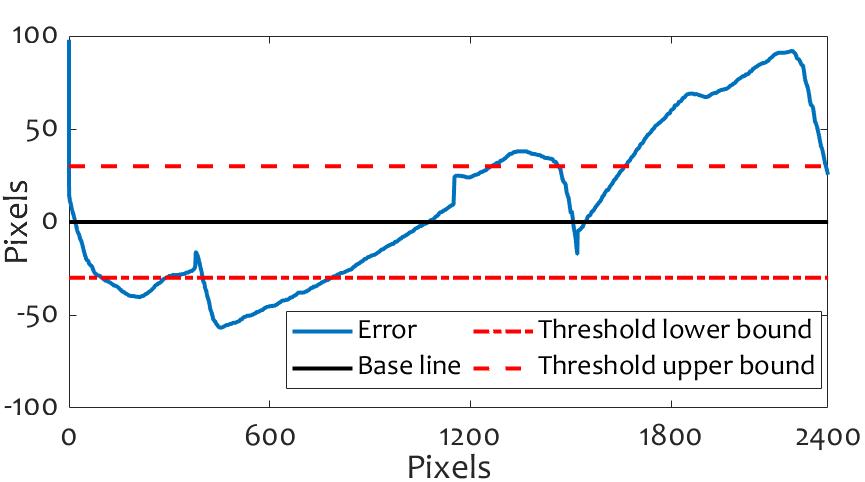}
		}
		\subfloat[trajectory error in virtual part]{
			\includegraphics[width=0.47\linewidth]{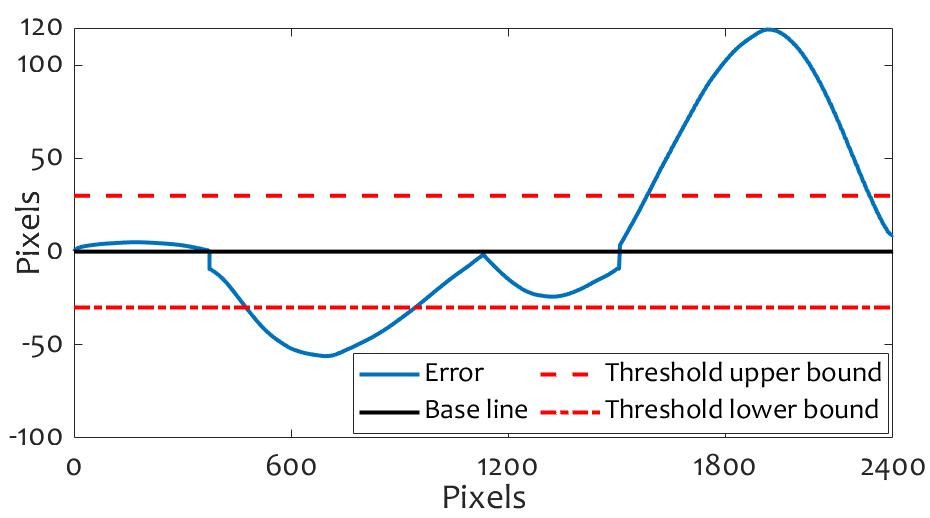}
		}
		\caption{Experimental results for physical and virtual testbed respectively without interaction}
		\label{fig:pic8}
		\vspace{-7pt}
	\end{figure}
The control protocol used in the experiment is the widely used PID control, which is shown as follow
\begin{equation}\label{PID}
    x(k + 1) = x(k) + {\varepsilon _p}({x_{target}} - x(k)),
\end{equation}
where $x(k)$ denotes the position of robot at time $k$, $\varepsilon_p$ is the coefficient of p-controller, $x_{target}$ is position of targets.
\subsection{Experiment A}
In the first part, we show the results of the experiment on the physical testbed and virtual testbed without interaction. The target positions on the physical testbed are set to [(500, 500) - (500, 900) - (1300, 900) -  (500, 900) -  (500, 500)], on the virtual testbed are set to [(500, 500) - (500, 900) - (1300, 900) -  (500, 900) -  (500, 500)] (the unit is pixel, and 1 pixel is about 2.5 mm). We also set a threshold of $\pm 30$ pixels allowable position deviation. The trajectory results of experiment are shown in Fig. \ref{fig:pic8} (a) and Fig. \ref{fig:pic8} (b). It should be noted that the actual trajectories are computed by real position data from physical testbed and virtual testbed. The trajectory errors are shown in Fig. \ref{fig:pic8} (c) and Fig. \ref{fig:pic8} (d). It can be seen that due to the slip and friction in physical and virtual testbeds, the actual trajectories have deviations from the target trajectories. The maximum errors exceed 91 pixels and 120 pixels in physical and virtual testbeds, respectively. Also, there exists a huge gap between the physical and virtual testbeds. The difference in slip and friction causes a difference in the dynamic models of the robot in two testbeds. Therefore, it is important to learning a dynamic model from the physical testbed, and mimic the robot movement in the virtual testbed, to construct a high-fidelity simulation.

\subsection{Experiment B}
\begin{figure}[t]
		\centering
		\subfloat[P-V trajectories]{
			\includegraphics[width=0.45\linewidth]{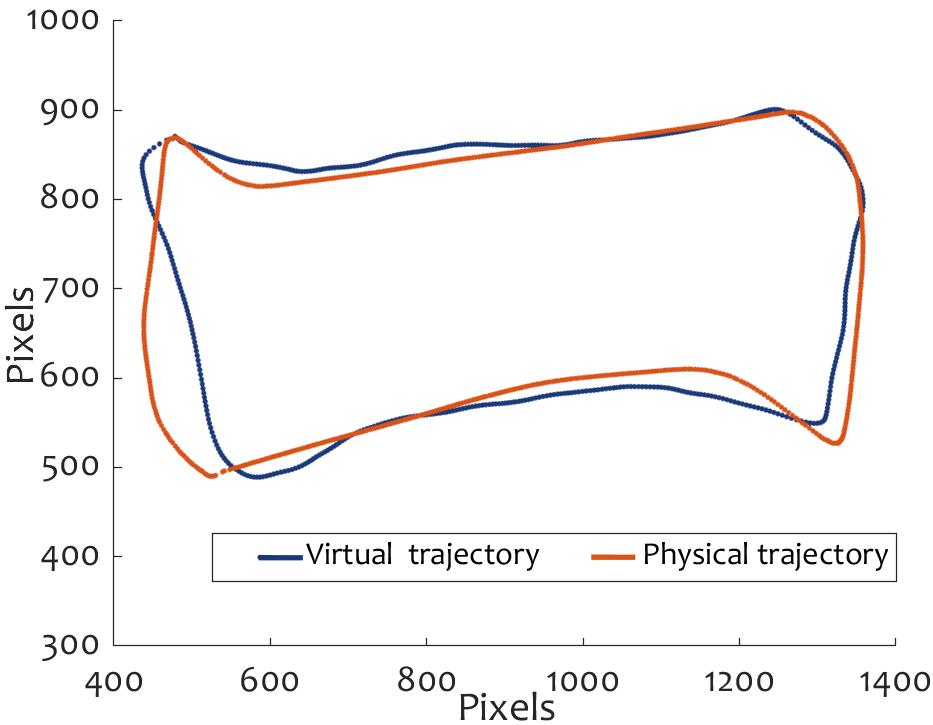}
			
		}
		\hspace{-5pt}
		\subfloat[P-V trajectory errors]{
			\includegraphics[width=0.48\linewidth]{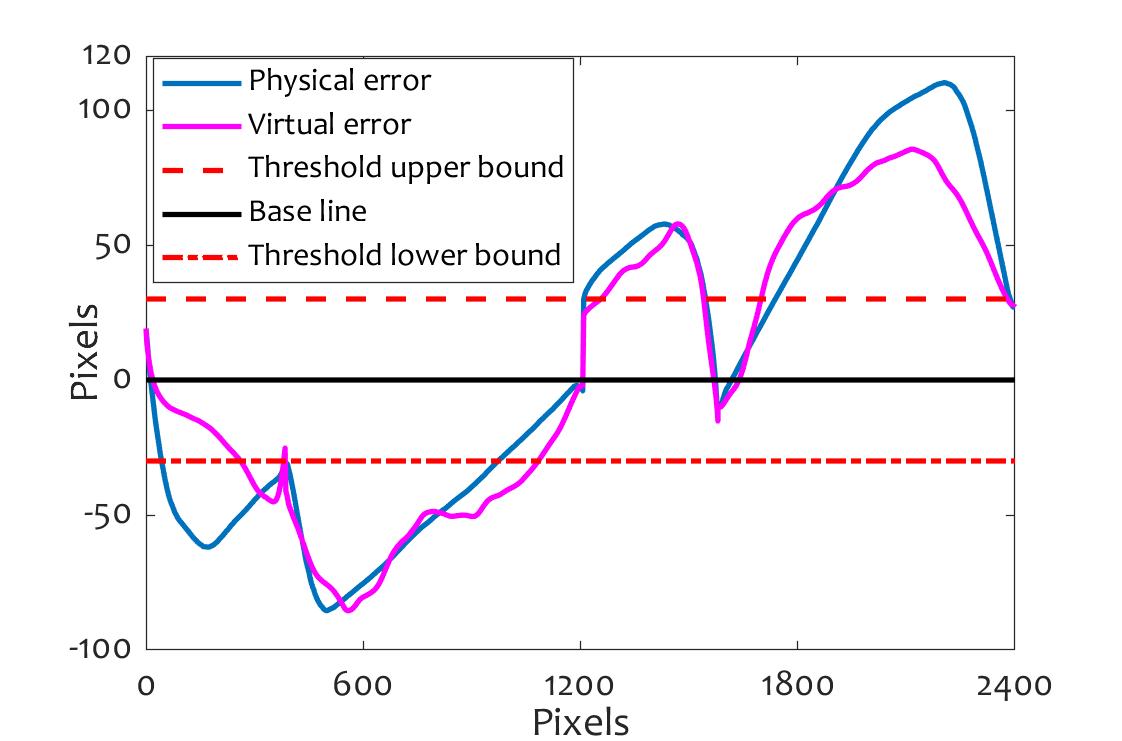}
		}
		
        \caption{Experimental results for physical and virtual testbed with learnt model}
		\label{fig:p-v}
	\end{figure}

The second part shows the experimental results on the physical and virtual testbed with part interaction, based on the data from physical testbed to virtual testbed to compute dynamics model learning process for the virtual model. The target position and the threshold remain the same. The results are shown in Fig. \ref{fig:p-v} (a) and Fig. \ref{fig:p-v} (b), respectively. 

		\begin{figure}[htb]
		\centering
		\subfloat[trajectory in first round]{
			\includegraphics[width=0.45\linewidth]{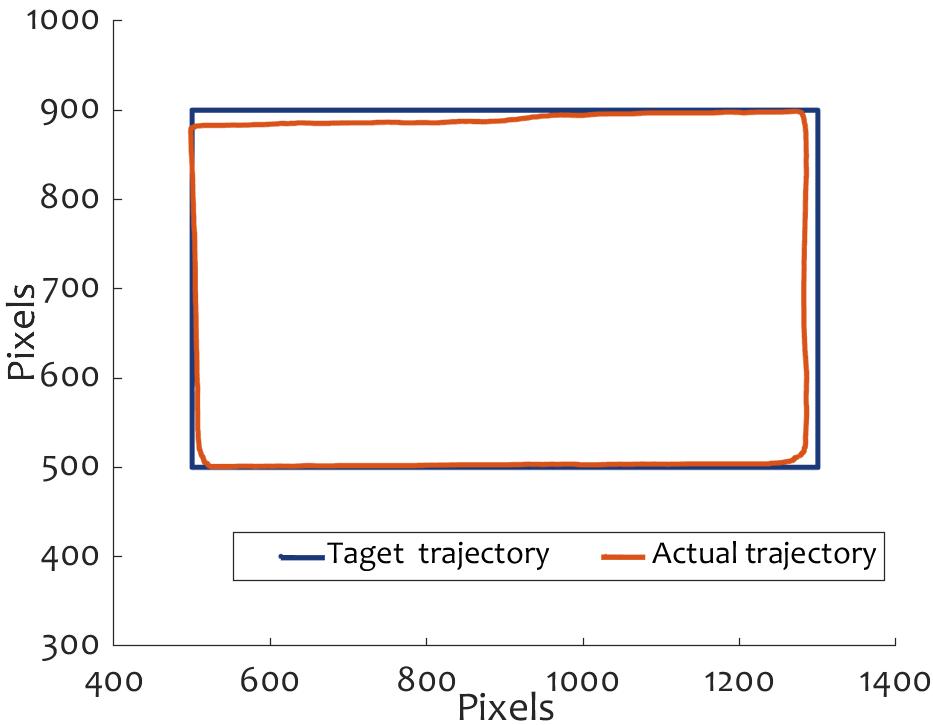}
		}
		\hspace{-5pt}
		\subfloat[trajectory in second round]{
			\includegraphics[width=0.45\linewidth]{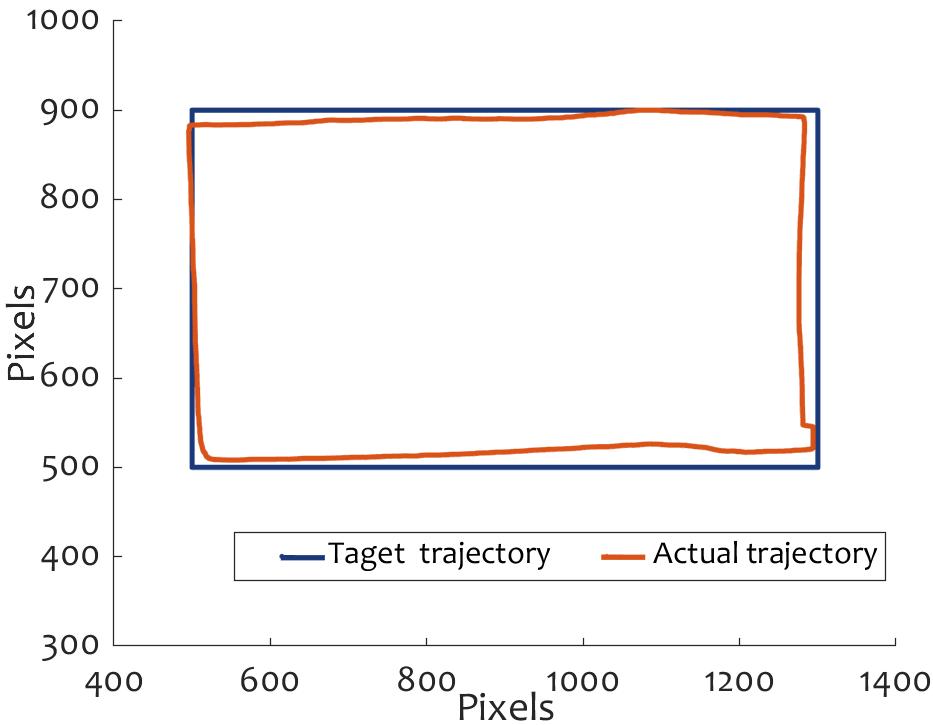}
		}
		
		\subfloat[trajectory error in first round]{
			\includegraphics[width=0.45\linewidth]{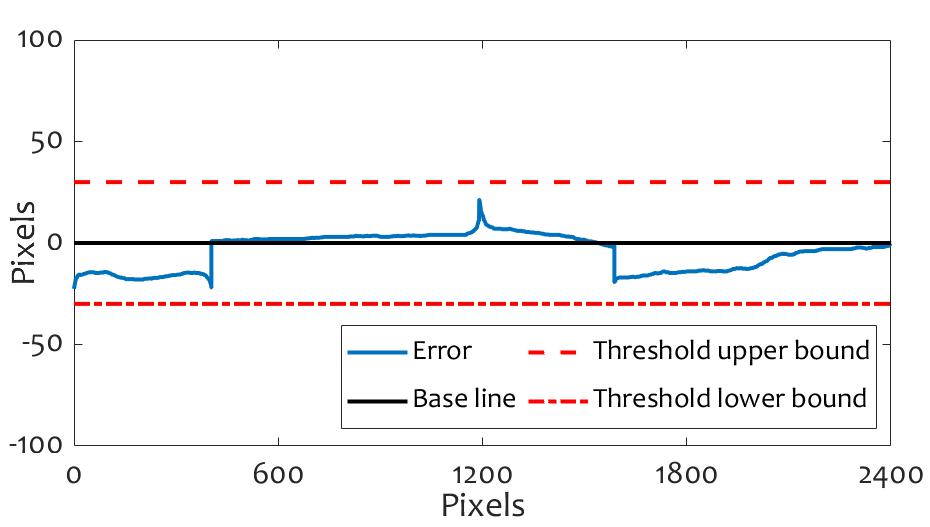}
		}
		\subfloat[trajectory error in second round]{
			\includegraphics[width=0.45\linewidth]{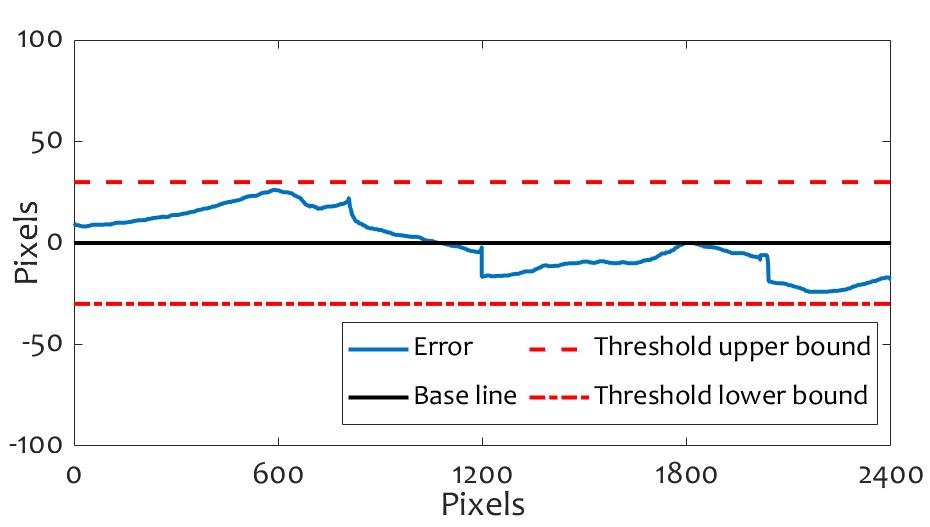}
		}
		\caption{Experimental results for physical testbed with fully P-V interaction}
		\label{fig:pic10}
		\vspace{-10 pt}
	\end{figure}
It is observed that after the training process, these two trajectories and error curves are close to each other, which means that the virtual model is similar to the physical model by introducing the learning procedure. The influence of slip and friction in the physical testbed can be introduced to the virtual testbed based on the procedure.

\subsection{Experiment C}
We show the experimental results on the physical testbed and virtual testbed with full interaction in the last part. In addition to the virtual testbed training model, we also utilize the online training process for the physical controller. 
The virtual testbed can compute the optimal feedback for the physical testbed control. The target position and the threshold remain the same. 
Two repeat experiments are conducted to show the effectiveness of the P-V interaction for control performance improvement.
The trajectory results of experiment are shown in Fig. \ref{fig:pic10} (a) and Fig. \ref{fig:pic10} (b). 
The trajectory errors are shown in Fig. \ref{fig:pic10} (c) and Fig. \ref{fig:pic10} (d).

It can be seen that the physical testbed gets a much better trajectory compared with the last two experiments. The actual trajectories are close to the target. The error stays in the threshold range all the time. Compared with Experiment A, the trajectory accuracy of robot movement improves about 300$\%$ with the feedback of the physical and virtual interaction.  The virtual testbed predicts the robot's movement in the physical testbed based on the learned model and further guides the robot to move better.

\section{CONCLUSION}
In this paper, we developed a novel  virtual-physical interactive mobile robotics testbed, which is named Robopheus. 
The system is compatible with heterogeneous kinematics robots and various control algorithms, providing a reliable verification environment for theory research. The operation data during physical experiments are further utilized for the dynamics model learning process.
The Robopheus can be run in physical, virtual, and physical-virtual interactive forms independently. 
The physical form provides a verification environment for heterogeneous kinematics robot control algorithms.
Furthermore, the operation data during the experiment can be stored in a database for the dynamics model learning process. 
The virtual form is operated based on the virtual models to perform high-fidelity simulation, which can be installed in computers and used anytime. 
The physical-virtual form provides the dynamics model learning process for the virtual model and online optimization of the physical testbed controller. 

The system can be used in various robotics applications, e.g., robot swarm, SLAM, heterogeneous robot cooperation. Users can easily construct these scenes in a virtual testbed before experimenting on physical ones. Also, the virtual-physical interaction can help with optimizing the performance of those applications. 
 
\balance
\bibliographystyle{ieeetr}
\bibliography{icra2021.bib}
\end{document}